\documentclass[10pt,twocolumn,letterpaper]{article}

\usepackage[pagenumbers]{cvpr} 

\definecolor{cvprblue}{rgb}{0.21,0.49,0.74}
\usepackage[pagebackref,breaklinks,colorlinks,allcolors=cvprblue]{hyperref}
\usepackage{array}

\def\confName{CVPR}
\def\confYear{2025}

\title{Unified Dense Prediction of Video Diffusion}

\author{
Lehan Yang$^{1}$~~~
Lu Qi$^{2}$~~~
Xiangtai Li$^{3}$~~~
Sheng Li$^1$~~~ 
Varun Jampani$^4$ ~~~
Ming-Hsuan Yang$^{2}$ ~~~
\\[0.2cm]
$^1$University of Virginia~~
$^2$University of California, Merced~~ \\
$^3$Nanyang Technological University~~ 
$^4$Stability AI~~
}

\newcommand{\ourmethod}{UDPDiff}
\newcommand{\newcolormap}{Pixelplanes}
\newcommand{\datasetname}{Panda-Dense}

\begin{document}
\maketitle
\begin{abstract}

We present a unified network for simultaneously generating videos and their corresponding entity segmentation and depth maps from text prompts.
We utilize colormap to represent entity masks and depth maps, tightly integrating dense prediction with RGB video generation. 
Introducing dense prediction information improves video generation's consistency and motion smoothness without increasing computational costs. 
Incorporating learnable task embeddings brings multiple dense prediction tasks into a single model, enhancing flexibility and further boosting performance. 
We further propose a large-scale dense prediction video dataset~\datasetname, addressing the issue that existing datasets do not concurrently contain captions, videos, segmentation, or depth maps. 
Comprehensive experiments demonstrate the high efficiency of our method, surpassing the state-of-the-art in terms of video quality, consistency, and motion smoothness.

\let\thefootnote\relax\footnote{\scriptsize{Correspondence to Lu Qi and Xiangtai Li.}}

\end{abstract}    
\section{Introduction}
\label{sec:intro}

\begin{figure}[tp]
    \centering
    \includegraphics[width=0.45\textwidth]{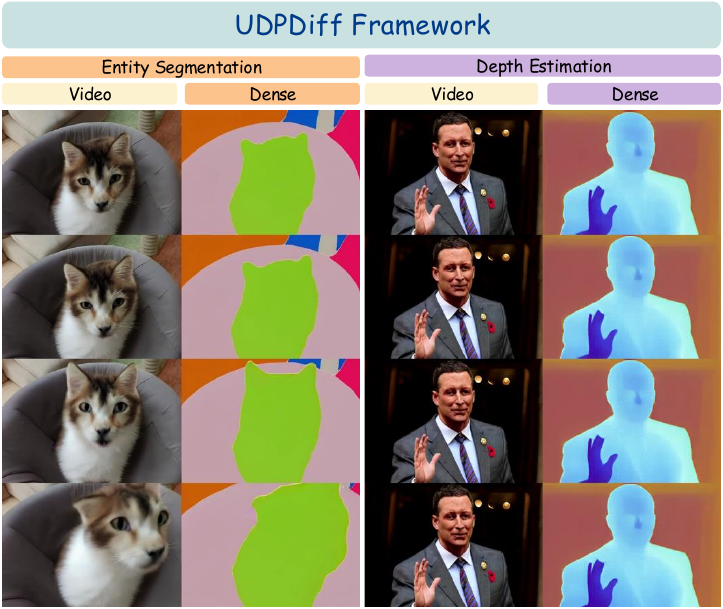}
    \caption{\textbf{Visualization results of multi-task~\ourmethod~model on image generation and dense prediction.} Our model can generate the video and the corresponding dense estimation. We incorporate two tasks in one multi-task model, including video entity segmentation and video depth estimation. Both segmentation and depth map have been encoded into RGB format as a video sequence, using~\newcolormap.}
    \label{fig.teaser}
    \vspace{-5mm}
\end{figure}

Video generation has witnessed substantial progress through diffusion models~\cite{ho2022video,singer2022make,chen2023videocrafter1,chen2024videocrafter2,ho2022imagen,wang2023modelscope,zhou2022magicvideo} and auto-regressive models~\cite{kondratyuk2023videopoet,sun2023emu,sun2024generative,wang2024emu3}. 
The main goal is to create coherent video clips from text prompts, facilitating diverse applications like video editing~\cite{feng2024ccedit,ceylan2023pix2video} and production~\cite{ge2024seedx,yang2024seedstory}.
These video generation techniques have a wide range of use cases, ranging from creative content generation to immersive multimedia experiences.

While existing foundational models can generate coherent video clips, they often face consistency issues, such as visual misalignment between frames and discrepancies with text prompts, which hinder some downstream editing tasks. Several approaches~\cite{zhao2024cv,yang2024cogvideox,esser2024scaling} have focused on effective module structures, such as 3D VAEs and MM-DiT blocks, for various vision tasks.  
However, these models still struggle to learn robust representations due to the complexity of the video generation task.
The work of REPA~\cite{yu2024representation} suggests that diffusion models can achieve faster and improved performance by aligning the representation with self-supervised methods. 
Nevertheless, these representations remain implicit without explicit semantic or geometry reasoning. 
An effective training signal should interpret generated results and integrate seamlessly with diffusion models. 
This insight motivates incorporating explicit signals into video diffusion training to enhance consistency, smoothness, and realism. 
Inspired by the unified representations in Emu Edit~\cite{sheynin2024emu} and UniGS~\cite{qi2024unigs}, we utilize two dense prediction signals: entity segmentation and depth estimation, both of which can be encoded as a unified representation, as shown in Fig.~\ref{fig.teaser}.
Entity segmentation ensures realistic object shapes and adherence to motion laws, while depth estimation enhances the model's perception of depth, form, and positional differences, enabling more structured and context-aware generation.

The first issue we tackle in this work is the lack of joint video, segmentation, and depth datasets. 
For this, we curate a video subset from the public Panda-70M~\cite{chen2024panda} by a series of data filtering such as video quality assessment and motion score.
To address the scarcity of data for video entity segmentation, we have developed a new pipeline that initiates with the EntitySeg~\cite{qilu2023high} to segment the initial frame, ensuring consistent granularity, and then employs the SAM2~\cite{ravi2024sam} video segmentation framework to extend the segmentation through all the video frames.
To obtain the depth data, we utilize the DepthCrafter~\cite{hu2024-DepthCrafter}, achieving consistent video depth estimation. 
Additionally, we have enhanced the existing captions using the Video-LLaVA~\cite{lin2023video}, providing more detailed and comprehensive descriptions, and improving the alignment from text to video.

The second challenge we address is designing a unified representation and learning architecture for jointly predicting different signals of RGB video and their corresponding frame segmentation and depth maps.
Compared to the use of colormaps~\cite{qi2024unigs} in image-level diffusion, to the use of colormaps for video-level diffusion presents additional complexities as it is necessary to adapt to multiple frames and consider the moving entities. Video-level representation is more challenging in two aspects, including the lack of large-scale training data wrapped with high-quality segmentation or depth annotations, as well as the lack of any unified %
architectural designs for multi-task training.

In this work, we introduce a~\newcolormap ~method to represent entity segmentation maps effectively. 
To better represent moving objects with varying locations and dense entities, we have randomized different colors to represent different entities. 
This approach also eliminates the potential ambiguity of predefined location-to-color mappings, allowing for less biased training.

We call our technique \ourmethod. 
Experimental results demonstrate that our method, benefiting from the assistance of segmentation and depth estimation, possesses better video synthesis quality and enhanced consistency, whether in the foreground or background. 
Additionally, our motion is more natural and smooth. 
The consistency and motion smoothness enhancements will provide stronger support and potential for downstream tasks such as video editing.
The main contributions of this work are as follows:
\begin{itemize}
    \item We introduce \ourmethod, the first adopted latent colormap representation in video generation and dense prediction, leveraging unified segmentation and depth estimation simultaneously with video generation.

    \item  We introduce a new unified representation~\newcolormap~ designed for video segmentation, removes the ambiguity of location-to-color mapping, and supports more dense entities. It also unifies the encoding of the depth estimation map, featuring a multi-task unified representation.

    \item We propose a new large-scale video dataset equipped with detailed captions, video segmentation, and depth estimation, powering the training of large-scale video diffusion dense prediction.

    \item Extensive experiments have demonstrated improvements in video quality, consistency, and motion smoothness in video generation from \ourmethod, with joint training.  
    Our work can inspire the use of foundation models for dense prediction tasks in the video itself and enhance downstream tasks' capabilities.

\end{itemize}

\section{Related Work}
\label{sec:related}
\vspace{1mm}

\noindent \textbf{Diffusion Model for Generation.}
Advances in latent diffusion models~\cite{rombach2022high} have contributed significantly to generation fields like text-to-image~\cite{rombach2022high,podell2023sdxl,peebles2023scalable} and video generation~\cite{ho2022video,singer2022make,ho2022imagen,wang2023modelscope}. Building on these pre-trained base models, techniques such as ControlNet~\cite{zhang2023adding} and LoRA~\cite{hu2021lora} enable more practical generation and editing applications~\cite{anydoor,yi2024diffusion,customdiffusion,chen2023videocrafter1,chen2024videocrafter2,zhou2022magicvideo} by incorporating conditional signals, such as segmentation layouts~\cite{wang2024ms,qi2024unigs}, depth/motion information~\cite{lapid2023gd,liang2025movideo}, or reference images/videos~\cite{anydoor,shi2024relationbooth}. Unlike those works, we focus on developing a unified model that simultaneously generates visual content and dense predictions, allowing both tasks to enhance each other.

\vspace{1mm}
\noindent \textbf{Diffusion Model for Dense Prediction.}
Dense prediction tasks like segmentation and depth estimation are fundamental in computer vision. While existing methods~\cite{long2015fully,chen2017deeplab,zhao2017pyramid,he2017mask,liu2018path,qi2019amodal,qi2020pointins,qi2024generalizable} mainly rely on discriminative models, diffusion-based approaches~\cite{qi2024unigs,xie2024zippo,wang2024semflow,ke2024repurposing,duan2023diffusiondepth,patni2024ecodepth,li2023open} have demonstrated state-of-the-art results. 
For instance, Dataset Diffusion~\cite{nguyen2024dataset} synthesis pixel-level semantic segmentation, and  InstructDiffusion~\cite{geng2024instructdiffusion} performs multi-task segmentation tasks. 
SemFlow~\cite{wang2024semflow} and UniGS~\cite{qi2024unigs} achieve comparable performance in semantic and entity segmentation~\cite{qi2021open,qi2022fine}.
Latent Diffusion Segmentation~\cite{van2024simple} creates a new mask VAE for panoptic segmentation, while Marigold~\cite{ke2024repurposing} surpasses prior arts in monocular depth estimation. 
In contrast to prior work focusing only on the accuracy of a single dense prediction task, we enhance both generation and downstream editing by jointly training multiple tasks. 

\begin{figure*}[t!]
    \centering
    \includegraphics[width=0.99\textwidth]{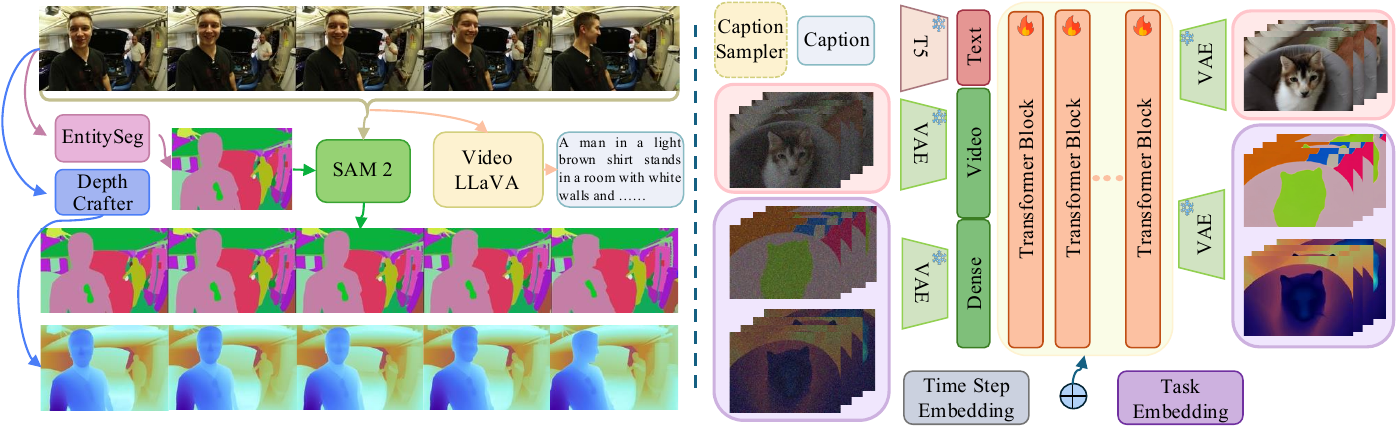}
    \caption{\textbf{Overview of the~\datasetname~ pipeline and the~\ourmethod~ framework.} \textbf{Left:} For segmentation, we use the first frame's results from EntitySeg as a prompt for SAM2, which then performs video segmentation across the entire sequence.  For depth estimation, DepthCrafter is used to generate video depth maps. For long prompts, Video LLaVA is used for captioning. \textbf{Right:} Similar to CogVideoX, our method~\ourmethod~ denoises the feature sequence in the latent space, encoding and decoding the latent using a 3D VAE. Video generation and dense prediction share a similar paradigm, using the same VAE for encoding and decoding through a unified representation. Task embeddings are applied to the time step embeddings, enabling more powerful differentiation of various tasks under a multi-task joint training model.}
    \label{fig.framwork}
    \vspace{-2mm}
\end{figure*}

\vspace{1mm}
\noindent \textbf{Unified Representation.}
The unified representation~\cite{sun2023emu,sun2024generative,wang2024emu3,qi2024unigs,tian2024mm,dong2023dreamllm,ge2024seedx} of generation and dense prediction is an emerging topic. Existing work generally follows two main pipelines, including auto-regressive-based and diffusion-based methods. 
Auto-regressive models~\cite{sun2023emu,sun2024generative,wang2024emu3,tian2024mm,dong2023dreamllm,ge2024seedx}, such as the Emu series~\cite{sun2023emu,sun2024generative,wang2024emu3}, predict image, video, or text tokens sequentially.
In contrast, diffusion-based methods~\cite{qi2024unigs,wang2024semflow,ke2024repurposing} treat dense prediction as a colormap that can be directly diffused. Our work focuses on the diffusion process, leveraging state-of-the-art performance demonstrated in video generation. 
Unlike existing diffusion-based approaches that primarily target image-level generation, our study is the first to introduce a unified diffusion model for video-level generation and dense prediction.

\section{Method}
\label{sec:method}

Based on CogVideoX~\cite{yang2024cogvideox}, we propose the~\ourmethod~framework, which aims to denoise images and dense predictions simultaneously based on text prompts. 
On the right side of Fig.~\ref{fig.framwork}, we showcase the pipeline of~\ourmethod ~that allows the integration of multiple tasks within the same model.
%
It guides the model in completing different generation tasks by inputting various task IDs through a learned task embedding.

\subsection{Dense Prediction Dataset}
\label{sec:method_dense_prediction_dataset}

There exist no readily available video datasets with entity segmentation and depth annotations.
We propose a large-scale captioned dense prediction video dataset~\datasetname, as shown in the left part of Fig.~\ref{fig.framwork}, including video entity segmentation and depth maps.
Based on the Panda-70M dataset, we randomly sampled a subset of approximately 300K samples, using a 13B Video-LLaVA for re-captioning. 
We retain the original brief captions from Panda-70M and the new detailed captions.

\vspace{1mm}
\noindent\textbf{Entity Segmentation.}
We leverage the image entity segmentation model of EntitySeg CropFormer~\cite{qilu2023high} coupled with the powerful SAM2~\cite{ravi2024sam} video segmentation to construct the video segmentation dataset.
We need to use the two models as the SAM model~\cite{kirillov2023segment} has inconsistent granularity issues, where the point grid initialization method can lead to overly fine or coarse segmentation maps. 
In particular, we extract the first frame for each video and use the EntitySeg CropFormer~\cite{qilu2023high} model for entity segmentation. 
We then use this segmentation mask as the prompt input for SAM2 to complete the propagation throughout the entire video frames. 
We also post-process these frames to remove small holes and the empty area.
Our segmentation dataset exhibits diversity in the number of entities, which can %
provide a rich learning signal to generate entities in both simple and complex scenes.

\vspace{1mm}
\noindent\textbf{Depth Estimation.} Video depth estimation has traditional methods based on single images, such as Depth Anything V2~\cite{yang2024depth}, which can accurately estimate the depth of each frame. 
However, linking these frames does not ensure consistency between frames, resulting in a series of flickers and noise, which is detrimental to training video diffusion. 
Another way is video-based diffusion methods, such as DepthCrafter~\cite{hu2024-DepthCrafter}, which can jointly process multiple video frames and
generate video depth maps with high precision and consistency. 
We use DepthCrafter to perform inference on a subset of our dataset and map the single-channel depth map to RGB space using a spectral-styled distance-based colormap projection.

\subsection{Encoding~\newcolormap}
\textbf{Review of Location Aware Colormap.} UniGS~\cite{qi2024unigs} colormap uses a pre-defined color grid with fixed size and colors. 
A location-based projection $\Psi$ from the entity centroid coordinates to the colors is used during encoding. 
This converts entity-level binary segmentation masks $M \in \{0, 1\}^{n\times h\times w}$ into an RGB colormap $M_c \in [0, 255]^{h\times w\times 3}$: $M_c = \Psi(M)$.

However, a manually designed color grid may not necessarily conform to the optimal distribution for model learning. 
Moreover, this color grid is fixed in size, which can lead to issues of different entities sharing the same color in densely populated multi-entity scenes. 
Furthermore, entities are likely in motion in video-dense prediction, making location-aware methods effective only in the first frame and creating ambiguities in subsequent frames.

\vspace{1mm}
\noindent\textbf{\newcolormap.}
To address these issues with the above colormap representation, we designed a new colormap that reduces ambiguity and is adaptable for dense multi-object segmentation. It is suitable for both video entity segmentation and video depth estimation. 
We introduce the~\newcolormap~as the unified representation. 
During the segmentation encoding, we use a random color generation function: ${ random :}\ [0,255]$, which produces a random integer in the range [0,255] representing a color channel value.

For all entity-level binary segmentation masks $M \in \{0, 1\}^{n\times h\times w}$, we randomly sample each of the RGB channels separately to obtain: $M_c = (r_n, g_n, b_n)$.
In this way, the color of each entity is independent and random, and we ensure that no duplicate colors appear on different entities, thereby enhancing the entity-discrimination capability of this representation. 
Freed from the constraints of location coordinates, entities' motion will not experience the color ambiguities caused by changes in location.

For depth maps, we use a spectral style value projection $\Upsilon$ to map depth $D \in [0, 1]^{h\times w}$ onto RGB spectral depth map $D_c \in [0, 255]^{h\times w\times 3}$ : $D_c = \Upsilon(D) $.
Compared to directly duplicating the depth channel into a three-channel RGB image, this color projection method adapts better to the real image distribution of VAE. It conforms more closely to the color representation of the diffusion prior. 
Single video data lacks explicit motion actions and hierarchical relationships, segmentation and depth estimation can abstract these motions, reducing implicitness, thereby expecting to achieve a representation that conforms more closely to the real-world physical laws of motion.

\subsection{Joint Generation Pipeline}

\vspace{1mm}
\noindent\textbf{Overview of CogVideoX.}
Our model is built upon CogVideoX 5B, which utilizes a 3D VAE for encoding the video sequence to a latent. 
In contrast to the 2D VAE used in Stable Diffusion, the 3D VAE can compress and encode latent temporally, offering better coherence and compressibility. 
Given an RGB video $\mathcal{X} \in \mathcal{R}^{T \times H \times W \times 3}$, where $T$, $H$ and $W$ are the frame number, height, and width of a video. 
The video is encoded by the 3D VAE into latent $z$. 
Compared to the UNet structure of diffusion models, Sora~\cite{videoworldsimulators2024} used a scalable transformer for long video generation. The transformer architecture scales more efficiently and has greater model capacity, making it more suitable for temporally involved video generation. 
Just as with image diffusion, as a Markov chain with forward and backward processes, the diffusion model gradually removes noise from a noisy video to obtain the final generated video.

Furthermore, video generation typically requires more detailed text prompts than image generation. 
The CLIP text encoder used in stable diffusion might not suffice for encoding long and complex prompts. 
Therefore, this model employs the more powerful T5~\cite{raffel2020exploring} model as its text encoder.

\vspace{1mm}
\noindent\textbf{\ourmethod~Framework.}
We input corresponding and more abstract video entity segmentation and depth estimation to enhance the guidance of dense prediction tasks on video generation tasks. 
This improves the explicitness of the input, enabling the model to understand scenes and motion better, thereby generating more realistic motion and consistent videos.
As shown in the right part of Fig.~\ref{fig.framwork}, we expanded the input layer and output layer, doubling the video input and output channels, and performed channel-wise concatenation of the video latent with one type of dense prediction latent as the input. 
The advantage of this approach is that it allows the guidance from dense prediction to affect all layers, thereby enhancing consistency.

Specifically, our input has four components: the latent code of the noised video $z_t^v$, the latent code of the unified dense colormap $z_t^c$, the text prompt $c_t$, and the task id $d$. 
Under the condition of the text prompt, we compute the task embedding from the given $d$, which serves as a joint condition. 
We use a transformer $f_\theta$ to denoise $z_0^v$ and $z_0^c$ to $z_t^v$ and $z_t^c$. 
During inference, $z_0^v$ and $z_0^c$ are encoded from random Gaussian noise. 
And through the channel-wise concatenation operation, the two groups of latents for video and dense prediction, $z_t^v$ and $z_t^c$, are concatenated into $z$, and at the t-th step represented as $z_t$, resulting in a total of 32-channel latent: $z_t = \text{CONCAT}(z_t^v, z_t^c)$.

\subsection{Learned Task Embedding}
To adapt multiple tasks using a single model, the intuitive method is to distinguish tasks using text prompts.
However, the text prompt condition based on semantics is an implicit task condition, which may not explicitly and correctly coordinate the target task sampling. 
Thus, this causes task semantic ambiguity and potentially misleads the guidance for dense prediction. 
Therefore, similar to the multi-task differentiation method used in Emu edit~\cite{sheynin2024emu}, we have effectively adapted it from image-level multi-task learning to video-level multi-task learning.
We employ a learnable task embedding $e_\theta^d$, which takes the task id $d$ as input and adds its output to the time step embedding $e_\theta^t(t)$ to obtain $t_d$:
\begin{equation}
\label{eq:task_emb}
t_d = e_\theta^d(d) + e_\theta^t(t) 
\end{equation}
During training, we jointly optimize this embedding layer while optimizing the generative diffusion model. 
The objective can be described as:
\begin{equation}
\label{eq:diff_loss}
\mathcal{L}_\text{train} =  \frac{1}{2}|| f_\theta(z_t, t_d, c_t) - \epsilon ||^2.
\end{equation}
During inference, in addition to the context prompt guiding video generation, we input a task ID into the task embedding layer, with options for segmentation or depth estimation. 
The network will generate outputs according to the specified task ID.

\section{Experiments}
\label{sec:exp}
In this section, we begin by evaluating the performance of our UDPDiff on a single auxiliary perception task for video generation, including segmentation and depth estimation.
Then, we demonstrate the overall effects of multi-task training. 
We utilize FVD and three temporal metrics from VBench~\cite{huang2023vbench} to evaluate all video synthesis tasks to access the generation quality and consistency.
For subject consistency, we calculate the similarity between frames using DINO features to measure whether the subject appearance remains consistent. 
For background consistency, we use CLIP features to calculate the similarity between frames, assessing whether the background scene remains consistent. 
For motion smoothness, we employ motion priors to evaluate the smoothness of the generated motions, checking whether the motion is fluid and conforms to the physical laws of the real world.
In the ablation study, we examine the benefits of learned task embeddings and unified dense prediction encoding to show the effectiveness of our method.

\subsection{Experimental Settings}

For all experiments, we utilize our~\datasetname~dataset based on the Panda-70M~\cite{chen2024panda}, having about 300K samples in total. 
To enhance the model's prompt generalization capabilities, we employ 80\% Video-LLaVA generated captions and 20\% original captions from Panda-70M.
We initialize our model using the CogVideoX 5B text-to-video model weights, resampling the original channel weights for the newly added channels. 
Our experiments are conducted on 8 NVIDIA A100-80G GPUs, with a batch size of 1 per GPU, fine-tuning for 40000 steps at a learning rate of 1e-5.
We note that all experiments are fine-tuned by CogVideoX due to the limited computing sources.
This training scheme ensures a fair comparison, we fine-tuned the original CogVideoX under equivalent conditions, allowing the experiments to yield consistent and reliable conclusions.
During the evaluation, we randomly select 1024 samples from the previously unused Panda-70M dataset for evaluation using FVD and VBench.

\subsection{Single-task Training}

\begin{table}
  \centering
  \begin{tabular}{@{}lccc@{}}
    \toprule
    Model & SC ($\uparrow$) & BC ($\uparrow$) & MS ($\uparrow$)  \\
    \midrule
    
    CogVideoX 5B & 94.57 & 95.80 & 97.67  \\
    \ourmethod~(segmentation) & \underline{94.98} & \underline{95.92} & \textbf{98.62}  \\
    \ourmethod~(depth estimation) & \textbf{95.91} & \textbf{96.24} & \underline{98.10}  \\
    \bottomrule
  \end{tabular}
  \caption{\textbf{Evaluation of consistency and smoothness in single task models.} SC stands for subject consistency, BC stands for background consistency, and MS stands for motion smoothness.}
  \label{tab:single_model}
\end{table}

For video entity segmentation and depth estimation tasks, we individually train separate models for each, producing outputs consisting of a video and one corresponding dense prediction output. 
As shown in Tab.~\ref{tab:single_model}, our method surpasses the original CogVideoX 5B in single-model performance for both the model with segmentation and the model with depth estimation, excelling in subject consistency, background consistency, and motion smoothness.
Thus, it is clear that introducing guidance from segmentation or depth estimation alone can enhance consistency.

\subsection{Multi-Task Training}

\begin{table}
  \centering
  \begin{tabular}{@{}lcccc@{}}
    \toprule
    Model & SC ($\uparrow$) & BC ($\uparrow$) & MS ($\uparrow$) & FVD ($\downarrow$) \\
    \midrule
    CogVideoX 5B & 94.57 & \underline{95.80} & 97.67 & 343.92 \\
    \ourmethod~(seg) & \underline{95.21} & 95.69 & \underline{98.24} & \underline{316.76} \\
    \ourmethod~(depth) & \textbf{97.07} & \textbf{96.89} & \textbf{99.23} & \textbf{302.55} \\
    \bottomrule
  \end{tabular}
  \caption{\textbf{Evaluation of consistency and smoothness in multi-task models.} SC stands for subject consistency, BC stands for background consistency, MS stands for motion smoothness, FVD stands for Frechet Video Distance.}
  \label{tab:multi_model}
\end{table}

We evaluate the multi-task model using the same architecture design, identical text prompts, and random seeds but with different task IDs as inputs.
As shown in Table~\ref{tab:multi_model}, when comparing FVD, our model's inference on depth estimation and segmentation significantly outperforms CogVideoX, demonstrating superior generation quality.
Additionally, our model's inference on depth estimation outperforms CogVideoX regarding subject consistency, background consistency, and motion smoothness. Meanwhile, our model's inference on segmentation surpasses CogVideoX across most metrics, except background consistency, where the performance is comparable.
In addition, the jointly trained multi-task model outperforms the single-task model, especially for depth estimation, 
This improvement can be attributed to the complementary signals from the two dense prediction tasks: segmentation enhances the realistic object shapes and motion adherence, while depth estimation strengthens the perception of depth, form, and positional differences.

\vspace{1mm}
\noindent\textbf{Video Depth Estimation} 
We also verify the effectiveness of the depth maps generated by our model.
Since the output of our dense prediction is an RGB image, we conduct a reverse process, converting the RGB image back into a depth map according to the original depth degree-to-color projection.
Our model employs a text-to-video synthesis approach, but the generated dense predictions lack ground truth.
Therefore, we input the videos produced by our model into DepthCrater to generate a pseudo-depth ground truth and evaluate Delta accuracy and RMSE on these 1024 samples. 
Table~\ref{tab:depth_estimation} compares the results using Depth Anything V2 with ViT-L as backbone trained on 595K precise synthetic images and 62M pseudo-labeled real images. 
Considering our model only trained on 300K videos, which is significantly smaller than Depth Anything V2, our model achieves an acceptable level of accuracy.

\begin{table}
  \centering
  \begin{tabular}{@{}lccc@{}}
    \toprule
    Model & Data & $\delta_1$ ($\uparrow$) & RMSE ($\downarrow$) \\
    \midrule
    Depth Anything V2 & 62.6 M & 0.5808 & 0.0907 \\
    \ourmethod~(multi-depth) & 300 K & 0.4176 & 0.1634 \\
    \bottomrule
  \end{tabular}
  \caption{\textbf{Evaluation of depth estimation.} $\delta_1$ stands for the delta accuracy, and RMSE stands for the root mean square error.}
  \label{tab:depth_estimation}
\end{table}

\subsection{Qualitative Evaluation}

\begin{figure*}[tp]
    \centering
    \includegraphics[width=0.99\textwidth]{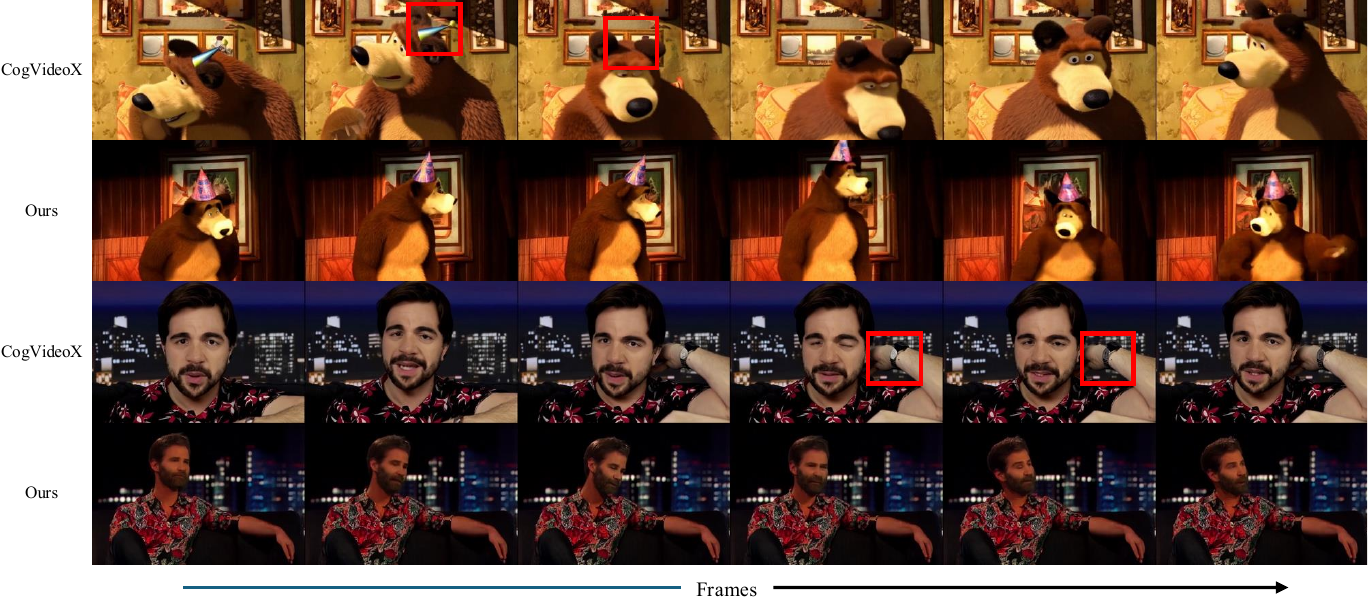}
    \caption{\textbf{Consistency qualitative comparison between CogVideoX and~\ourmethod}. Six frames are evenly sampled from the generated video, with the horizontal axis representing time and the frame index gradually increasing. Inconsistent parts are annotated with red bounding boxes, including disappearances, color changes, and shape changes.}
    \label{fig.cons}
\end{figure*}

\begin{figure*}[tp]
    \centering
    \includegraphics[width=0.99\textwidth]{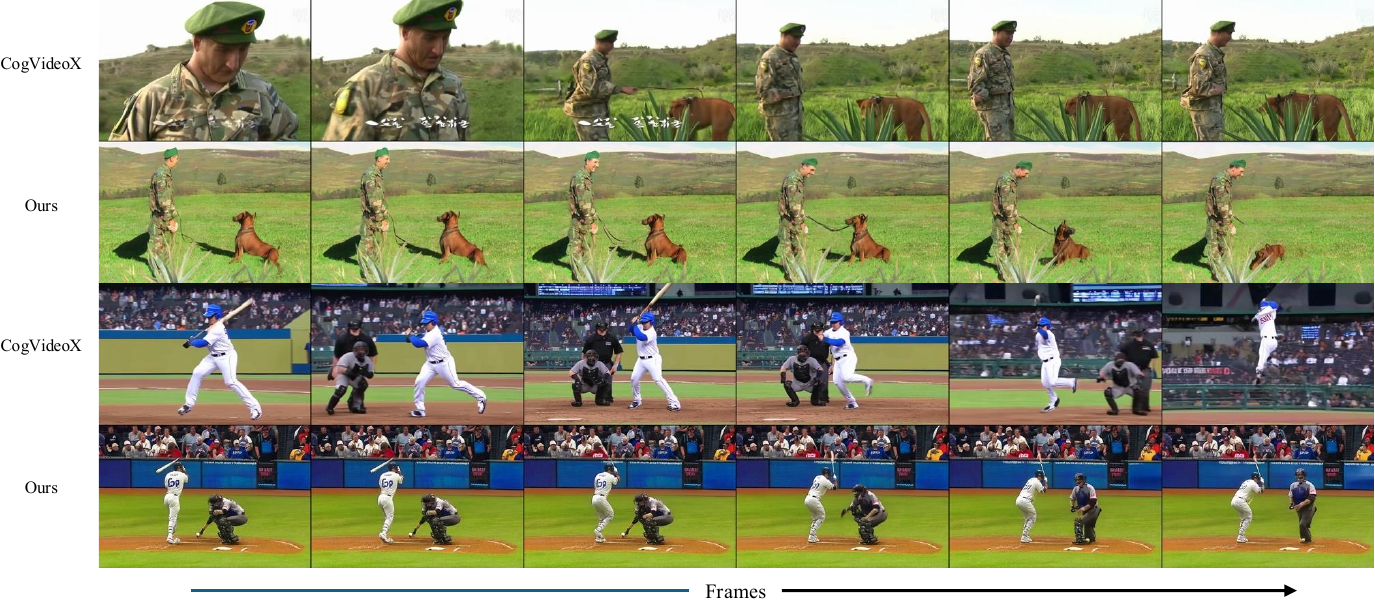}
    \caption{\textbf{Quality qualitative comparison between CogVideoX and~\ourmethod}. Six frames are evenly sampled from the generated video, with the horizontal axis representing time and the frame index gradually increasing. Our advantages are reflected in clearer and sharper entities, more realistic motion, and better generation of dense scenes.}
    \label{fig.quality}
\end{figure*}

\begin{figure*}[tp]
    \centering
    \includegraphics[width=0.99\textwidth]{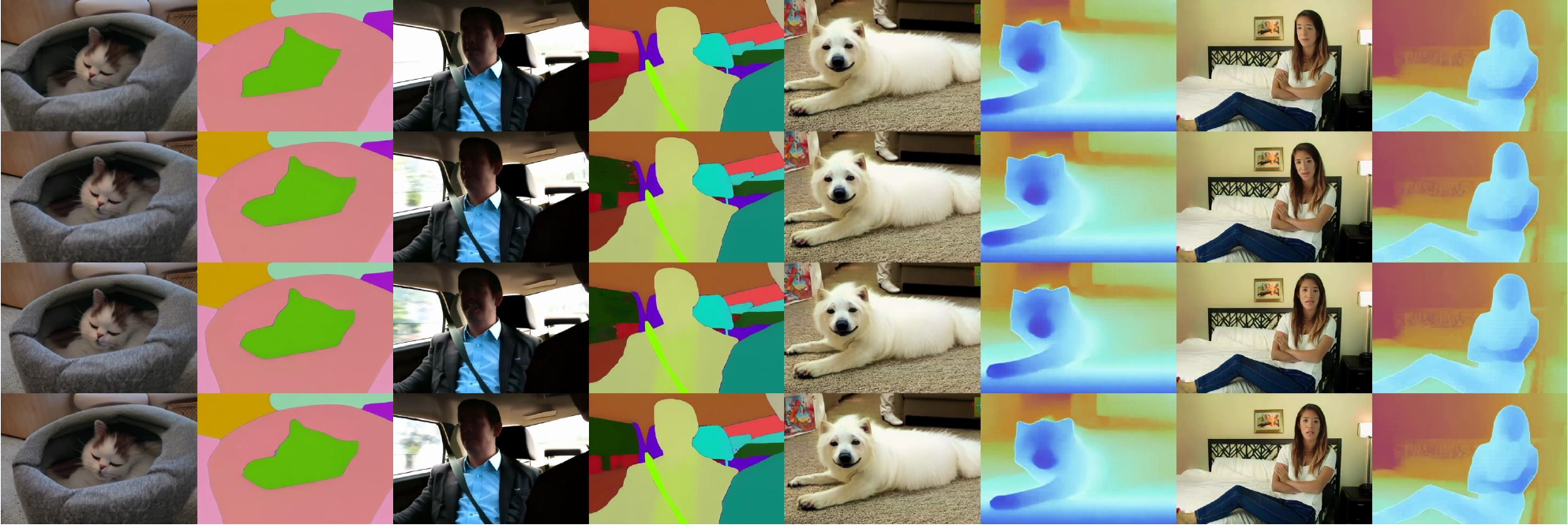}
    \caption{\textbf{Visualization of the video generation associated with dense prediction.} For each sample, we evenly sample four frames, with the left column of each sample representing the generated video and the right column representing the generated dense prediction. The left half of the samples is video entity segmentation, and the right half is video depth estimation. Our model can generate high-quality, dense predictions simultaneously with almost no increase in computational cost.}
    \label{fig.seg_depth}
\end{figure*}

We demonstrate the improvement of consistency in Fig.~\ref{fig.cons}. 
In the videos generated by CogVideoX, inconsistencies are annotated with red bounding boxes to highlight the issues.
For instance, the party hat disappears after several frames in the first example, and the watch's color changes in the second example.
In contrast, the videos generated by our method exhibited greater consistency and smoother motion.
At the same time, we demonstrate in Fig.~\ref{fig.quality} the higher quality of our video generation, where we can create more realistic motion trajectories for moving objects and motions that adhere more closely to the physical laws of the real world. 
Our method can generate clearer and sharper entities in dense scenes, such as those with multiple people or interactions among multiple individuals.
This all benefits from incorporating segmentation and depth estimation tasks, which enhance the video generation model's capabilities in scene perception and entity interaction.
In Fig.~\ref{fig.seg_depth}, we demonstrate the results of our method simultaneously generating video and corresponding dense predictions, including video entity segmentation and video depth estimation.
Dense predictions are represented by the unified representation~\newcolormap, directly generated by the diffusion model and decoded using the same video 3D VAE. 
Our model can achieve high-quality segmentation and depth estimation without significantly increasing computational cost, bringing about greater efficiency.

\subsection{Ablation Study}

\begin{table}
  \centering
  \begin{tabular}{@{}lc@{}}
    \toprule
    Model & Running time(s) \\
    \midrule
    CogVideoX 5B & 204.46  \\
    Ours (single task) & 205.75 \\
    Ours (multi task) & 205.81 \\
    \bottomrule
  \end{tabular}
  \caption{\textbf{Model inference speed comparison.} The inference speed of our single-task and multi-task models is compared with CogVideoX 5B, measured in seconds.}
  \label{tab:model_running_time}
\end{table}

\begin{table}
  \centering
  \begin{tabular}{@{}lcccc@{}}
    \toprule
    Method & SC ($\uparrow$) & BC ($\uparrow$) & MS ($\uparrow$) & FVD ($\downarrow$) \\
    \midrule
    Text prompt & 95.17 & 95.78 & 98.67 & 321.43 \\
    Task embedding & \textbf{97.07} & \textbf{96.89} & \textbf{99.23} & \textbf{302.55} \\
    \bottomrule
  \end{tabular}
  \caption{\textbf{Ablation study on task embedding}, SC stands for subject consistency, BC stands for background consistency, MS stands for motion smoothness, FVD stands for Frechet Video Distance.}
  \label{tab:ablation_task_embed}
\end{table}

\begin{table}
  \centering
  \begin{tabular}{@{}lccc@{}}
    \toprule
    Method & SC ($\uparrow$) & BC ($\uparrow$) & MS ($\uparrow$) \\
    \midrule
    Location-aware colormap & 81.26 & 79.33 & 88.79 \\
    \newcolormap & \textbf{94.98} & \textbf{95.92} & \textbf{98.62} \\
    \bottomrule
  \end{tabular}
  \caption{\textbf{Effectiveness of~\newcolormap compared to UniGS's location-aware colormap.} SC stands for subject consistency, BC stands for background consistency, and MS stands for motion smoothness.}
  \label{tab:colormap_compare}
  \vspace{-5mm}
\end{table}

\vspace{1mm}
\noindent\textbf{Model Inference Speed}.
Specifically, our model only increases the input and output channels and introduces an additional task embedding without significantly increasing the number of parameters in the original network. 
Here, we compare the inference speed of the original CogVideoX 5B with our single-task model and the multi-task model with task embedding. 
On a single NVIDIA A100-80G, performing inference for 50 steps for a 49-frame video, all using the same text prompt, the original CogVideoX outputs videos, while our model outputs videos and dense predictions. 
As shown in Tab~\ref{tab:model_running_time}, our inference speeds are almost similar within the range of uncertainty. 
The added channels do not affect the inference speed, demonstrating the efficiency of our method.

\vspace{1mm}
\noindent\textbf{Learned Task Embedding}. We compare the video generation metrics of the multi-task model under the condition of inferring depth estimation, using both text prompts and task embeddings to distinguish between segmentation and depth estimation tasks. 
Tab.~\ref{tab:ablation_task_embed} shows the results under the task embedding method are improved in consistency, motion smoothness, and FVD compared to those using text prompts. 
This demonstrates the effectiveness of task embeddings in distinguishing different dense prediction tasks more effectively. It allows the model to understand the current task more explicitly and receive the correct guidance.

\vspace{1mm}
\noindent\textbf{Unified Dense Prediction Encoding}
To demonstrate that our~\newcolormap~is more suitable for video generation than the location-aware colormap proposed by UniGS, we follow the colormap encoding method mentioned in UniGS, projecting the centroid of an entity in the initial frame onto an RGB color and populating it into subsequent frames.
Tab.~\ref{tab:colormap_compare} shows
the proposed unified model performs better than the single-task method using a random dense prediction encoding scheme, 
On the other hand, the colormap approach does not work as effectively, proving that our method eliminates the ambiguity of location changes and provides better guidance for multiple objects.

\begin{table}
  \centering
  \begin{tabular}{@{}lccc@{}}
    \toprule
    Partition & SC ($\uparrow$) & BC ($\uparrow$) & MS ($\uparrow$) \\
    \midrule
    (0.3, 0.7) & 95.89 & 96.54 & 98.16 \\
    (0.7, 0.3) & 96.72 & 96.37 & 96.97 \\
    (0.5, 0.5) & \textbf{97.07} & \textbf{96.89} & \textbf{99.23} \\
    \bottomrule
  \end{tabular}
  \caption{\textbf{Multi task data modality partition for segmentation and depth estimation.} In the partition column bracket, the first element indicates the data partition for segmentation, and the second indicates the data partition for depth estimation. SC stand for subject consistency, BC stand for background consistency, MS stand for motion smoothness.}
  \label{tab:multi_task_data_partition}
\end{table}

\begin{table}
  \centering
  \begin{tabular}{@{}lccc@{}}
    \toprule
    Method & SC ($\uparrow$) & BC ($\uparrow$) & MS ($\uparrow$) \\
    \midrule
    Depth Anything V2 & 94.87 & 96.29 & 96.13 \\
    DepthCrafter & \textbf{95.91} & \textbf{96.24} & \underline{98.10} \\
    \bottomrule
  \end{tabular}
  \caption{\textbf{Depth Estimation Method for ~\datasetname~ Pipeline.} Different depth data creation method, evaluate our model's performance on each dataset. SC stands for subject consistency, BC stands for background consistency, MS stands for motion smoothness.}
  \label{tab:depth_method}
      
\end{table}

\begin{table}
  \centering
  \begin{tabular}{@{}lccc@{}}
    \toprule
    Method & SC ($\uparrow$) & BC ($\uparrow$) & MS ($\uparrow$) \\
    \midrule
    8 $\times$ 8 & 93.12 & 93.73 & 96.18 \\
    12 $\times$ 12 & 92.96 & 93.59 & 96.35\\
    EntitySeg & \textbf{94.98} & \textbf{95.92} & \textbf{98.62} \\
    \bottomrule
  \end{tabular}
  \caption{\textbf{SAM 2 Init Prompt for ~\datasetname~ Pipeline.} Point grid initialization, compared to EntitySeg initialization, evaluate our model's performance on each dataset. SC stands for subject consistency, BC stands for background consistency, MS stands for motion smoothness.}
  \label{tab:seg_method}
\end{table}

\vspace{1mm}
\noindent\textbf{Multi Task Data Partition}
Considering the varying difficulty and guidance strength levels for segmentation and depth estimation tasks, we employ different partitions for segmentation and depth estimation during the joint training of the multi-task model. 
We ablate partition ratios of 0.3, 0.5, and 0.7 for each dense data type.
As shown in Tab~\ref{tab:multi_task_data_partition}, the best performance is observed when the data volumes for segmentation and depth estimation are equal.
The guidance provided by segmentation and depth estimation can be considered to be of a comparable degree.

\vspace{1mm}
\noindent\textbf{Dataset Pipeline}
For depth estimation, we compare the usage of Depth Anything V2 and DepthCrafter to generate two different depth datasets and train our single depth model on these datasets. After training on both datasets, as shown in Tab.~\ref{tab:depth_method}, we evaluate the video generation quality and find that DepthCrafter outperformed Depth Anything V2. This is because DepthCrafter natively supports video depth estimation, whereas Depth Anything V2 performs estimation frame by frame in independent frames. Such an image-level approach loses inter-frame consistency, potentially introducing additional noise into the training.

We employ a point grid as a prompt to intuitively automate SAM-2 video segmentation. As shown in Tab.~\ref{tab:seg_method}, we compare 8$\times$8 and 12$\times$12 point grids of SAM-2 with our EntitySeg initialization prompt. The observation is that inconsistencies in segmentation granularity exist and can affect the quality of video generation.

\section{Conclusion}
This paper introduces a novel and effective improvement in video generation by incorporating dense prediction tasks such as segmentation and depth estimation to guide higher-quality and consistent video generation. 
The key to our method is a redesigned color projection method that conforms to video encoding patterns. This method simultaneously generates the segmentation mask and depth map alongside the RGB video. 
Additionally, a large-scale captioned video dense prediction dataset is constructed with the proposed method. 
Extensive experiments prove the effectiveness of each of our modules, and the joint training of the multi-task model enhances efficiency and further improves consistency and quality.

{
    \small
    \bibliographystyle{ieeenat_fullname}
    \bibliography{main}
}


\appendix
\section{Appendix}

This supplementary material provides additional visualization results, details about the ~\datasetname~ video dense prediction dataset, and thorough descriptions of the training and inference processes for video generation and dense prediction, which includes segmentation and depth estimation. 
The contents of this supplementary material are organized as follows:

\begin{itemize}
    \item \datasetname~Dataset details.
    \item Detailed training and inference information for two tasks.
    \item More ablation studies on the model.
    \item Additional visualization results, including examples of video generation and the dense prediction.
\end{itemize}

Additionally, please see the mp4 file in our supplementary material to view the \underline{\textit{recorded video}} that provides a concise overview of our paper.

\vspace{1mm}
\section{\datasetname~Dataset Details}

\begin{figure}[t!]
    \centering
    \includegraphics[width=0.45\textwidth]{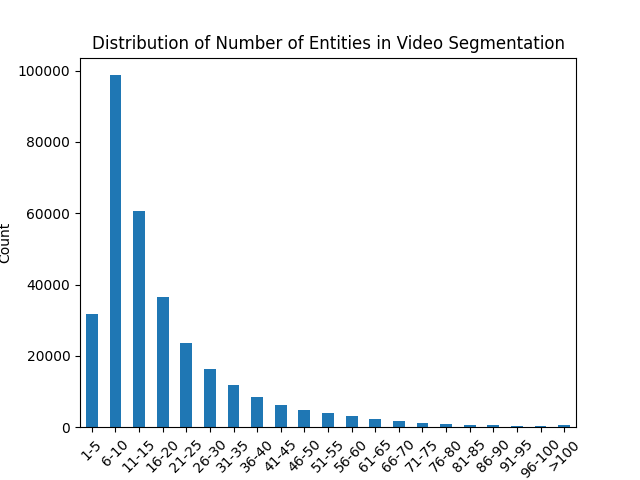}
    \caption{\textbf{Distribution of entity number in the video segmentation dataset.} The X-axis represents the groups of the total number of entities in a single video, while the Y-axis represents the total number of videos appearing in this group.}
    \label{fig.seg_stat}
\end{figure}

In Fig.~\ref{fig.seg_stat}, we present the distribution of the number of entities in our video segmentation dataset. 
The data shows a broad distribution of entities, with most concentrated in the range of 1 to 20. 
Additionally, our dataset includes numerous dense segmentation samples that contain multiple entities in complex scenes.

\section{Training / Inference Details}
In our work, we train our model using the \datasetname~data set and initialize it with the weights of the CogVideoX 5B model. 
We duplicate the weights from the original 16 channels to accommodate the added input and output channels.

This adjustment modifies the original layers instead of adding extra branches, which would disrupt the existing weights. 
In other words, this set of weights cannot be expected to directly output two feasible and identical video sequences after adding channels, so complete fine-tuning is required.

\begin{figure}[t]
    \centering
    \includegraphics[width=0.45\textwidth]{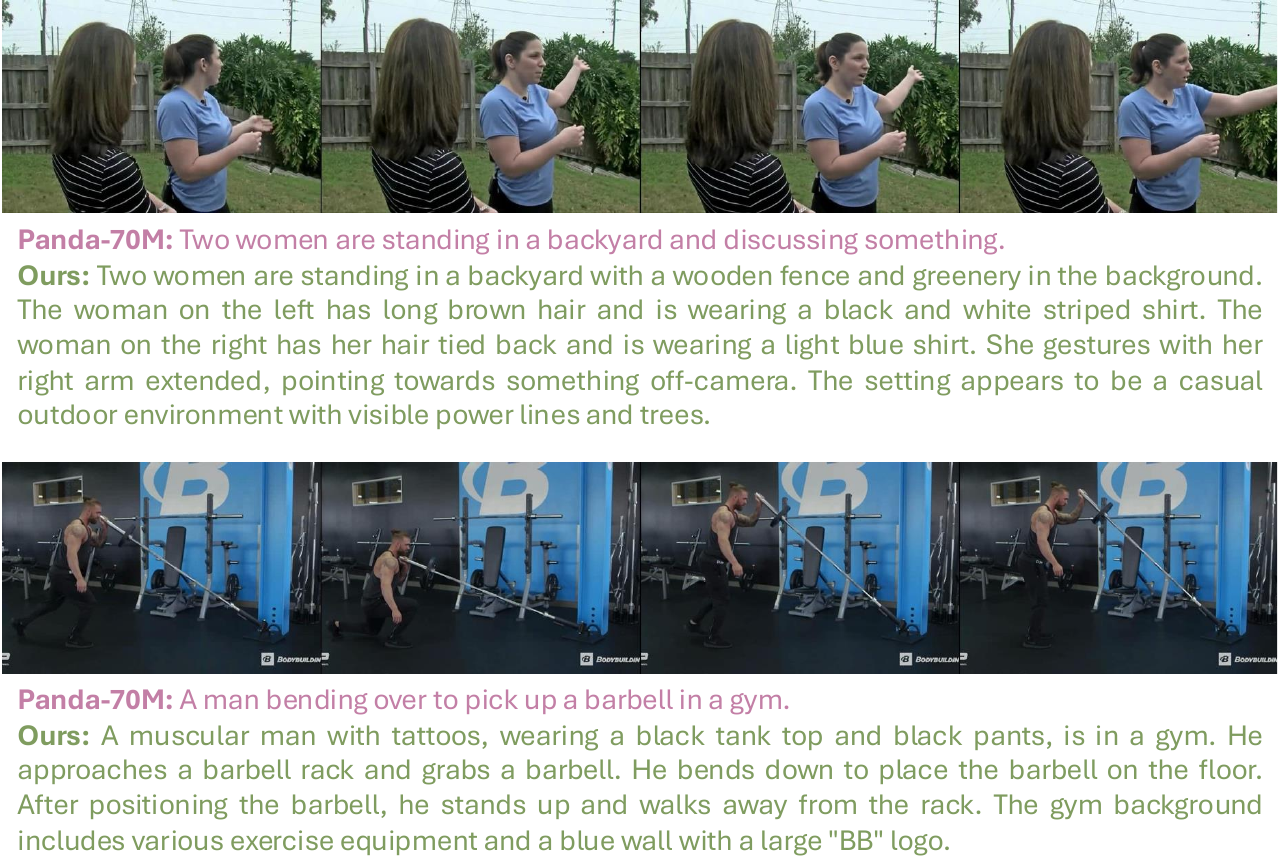}
    \caption{\textbf{Caption comparison.} Refer to the same video, compare the original Panda-70M caption and our~\datasetname~caption.}
    \label{fig.caption_compare}
\end{figure}

Regarding prompts, long captions offer more detailed descriptions and greater control, while short captions provide users with flexibility. 
We demonstrate the comparison between Panda-70M original captions and our~\datasetname~captions in Fig.~\ref{fig.caption_compare}.
To balance strong scene description capabilities with user input convenience, we utilize long captions from our dataset with a probability of 0.8 and short captions from Panda-70M with a probability of 0.2 for each video. 

Training is conducted under these conditions for 40000 steps. 
During inference, we employ DPM++ to execute 50 inference steps, generating 49 video frames along with their corresponding dense prediction maps.


\vspace{1mm}
\noindent\textbf{Video Depth Estimation.}
During inference, the video and the corresponding depth estimation map are produced, with the depth map appearing as an RGB value map similar to a visualization. 
Therefore, we should convert the depth map to a single-channel depth value between 0 and 1. 
Since the depth-to-color projection $P_d$ is a unidirectional function, we can not project the RGB colors back to depth values. To project it back, we sampled 256 RGB values with equal intervals in the 0-1 depth range, as: $P_c = P_d\left(\frac{k}{256}\right), \quad \text{for } k = 1, 2, \ldots, 256$. Thus, we have the discrete color-to-depth projection $P_c$.
For each pixel, we calculate the distance between its RGB value and the available discrete values in $P_c$, and then project back to the depth value.
After propagating the depth for all frames in the video, we obtain a distance-based video depth estimation map.

\vspace{1mm}
\noindent\textbf{Multi-task.}
The multi-task model defines a dictionary mapping different tasks to specific IDs, with segmentation assigned as 0 and depth estimation as 1. 
This ID serves as the input for the task embedding. 
During training, we randomly sample different tasks, corresponding to different task IDs and their respective dense prediction maps. 
During inference, in addition to the caption, a specific task ID must be provided as input.
The video channel consistently outputs video, while the dense prediction channel outputs either segmentation or depth estimation under the guidance of task embedding.

\section{Ablation Studies}

\begin{table}[t]
  \centering
  \begin{tabular}{@{}lccc@{}}
    \toprule
    Strategy & SC ($\uparrow$) & BC ($\uparrow$) & MS ($\uparrow$) \\
    \midrule
    Random initialization & 95.15 & 95.48 & 97.99 \\
    Duplicate initialization & \textbf{97.07} & \textbf{96.89} & \textbf{99.23} \\
    \bottomrule
  \end{tabular}
  \caption{\textbf{Ablation study on initialization strategy on additional channel weight.} Different additional channel weight initialization strategy for input and output layers. `SC', `BC' and `MS' stand for subject consistency, background consistency, and motion smoothness.}
  \label{tab:supp_channel_init}
  \vspace{1mm}
      
\end{table}

\begin{table}[t]
  \centering
  \begin{tabular}{@{}lcccc@{}}
    \toprule
    Strategy & Multi-task & SC ($\uparrow$) & BC ($\uparrow$) & MS ($\uparrow$) \\
    \midrule
    Scratch & - & 81.28 & 76.39 & 84.46 \\
    Scratch & \checkmark & 84.32 & 81.63 & 89.31 \\
    CogVideoX & \checkmark & \textbf{97.07} & \textbf{96.89} & \textbf{99.23} \\
    \bottomrule
  \end{tabular}
  \caption{\textbf{Ablation study on initialization strategy on whole model weight.} Different model-level weight initialization methods for~\ourmethod, comparing initialize from CogVideoX and train from statch. `SC', `BC' and `MS' stand for subject consistency, background consistency, and motion smoothness.}
  \label{tab:supp_weight_init}
      
\end{table}

\begin{figure}[t]
    \centering
    \includegraphics[width=0.47\textwidth]{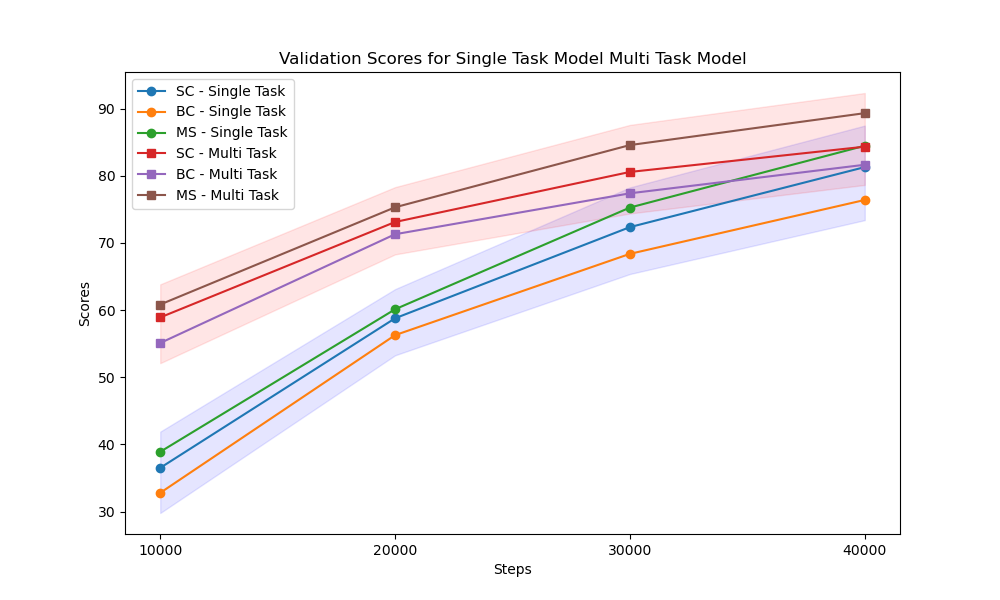}
    \caption{\textbf{Converge speed comparison.} Train from scratch, comparing single-task training and multi-task training converge speed. `SC', `BC' and `MS' stand for subject consistency, background consistency, and motion smoothness.}
    \label{fig.converge_comparison}
\end{figure}

\vspace{1mm}
\noindent\textbf{Additional Channel Weight Initialization Method.}
In our work, we augment the input and output layers with an additional 16 channels to accommodate the inputs and outputs of the dense prediction channels. 
Specifically, we duplicate the original 16 channels of the input and output layers of CogVideoX and concatenate the weights of the two sets of 16 channels to form 32-channel inputs and outputs.
We conduct ablation experiments using the multitask model inference depth estimation to explore different initialization methods for the weights of the additional channels, including using the pre-trained weights from CogVideoX and random initialization. 
As shown in Tab~\ref{tab:supp_channel_init}, utilizing the weights from CogVideoX's input and output layers as duplicate initialization outperforms random initialization. 
Although increasing the number of channels and concatenating them disrupts the original weight distribution, using pre-trained weights for dense prediction features still results in better adaptation performance.

\vspace{1mm}
\noindent\textbf{Pretraining.}
Since our method is based on CogVideoX 5B, we directly utilize the weights of CogVideoX 5B's text-to-video model to initialize all layers except for the input and output layers. 
For the input and output layers, we continue to adopt the duplicating channel approach by concatenating the original weights of CogVideoX.
In this experiment, we compare this strategy with an entirely random initialized multi-task model, both trained for 40000 steps. 
As shown in Tab~\ref{tab:supp_weight_init}, the method initialized with CogVideoX outperforms the training from scratch. 
This is because CogVideoX was trained on a dataset much larger than ours and employed various training tricks. 
Our fine-tuning approach cannot achieve good results on datasets of this scale.
Using pre-trained weights allows the model to learn from an appropriate distribution, resulting in better performance.
To further illustrate the advantages of our multi-task training strategy beyond the fine-tuning stage, we compare the performance of a single-task model trained from scratch with that of a multi-task model, as shown in Tab.~\ref{tab:supp_weight_init}.
Multi-task training delivers better performance under the same conditions.
Additionally, we present in Fig.~\ref{fig.converge_comparison} the validation scores of the single-task and multi-task models at different steps. 
The multi-task model achieves higher scores at each step and exhibits a smoother increase in the later stages, converging more rapidly. 
Therefore, our multi-task training strategy also accelerates the training process.

\begin{table}[t]
  \centering
  \begin{tabular}{@{}lccc@{}}
    \toprule
    Model Size & SC ($\uparrow$) & BC ($\uparrow$) & MS ($\uparrow$) \\
    \midrule
    2B & 87.29 & 89.19 & 91.76 \\
    5B & \textbf{94.98} & \textbf{95.92} & \textbf{98.62} \\
    \bottomrule
  \end{tabular}
  \caption{\textbf{Ablation study on model size.} Compare the performance of our method using CogVideoX at 2B and 5B model sizes. `SC', `BC' and `MS' stand for subject consistency, background consistency, and motion smoothness.}
  \label{tab:supp_model_size}
      
\end{table}

\vspace{1mm}
\noindent\textbf{Model Parameter.}
The original CogVideoX model is available in 2B and 5B. 
Our initial experiments utilize the 2B model.
Although the 2B model is more efficient, it has a smaller capacity, which may result in insufficient performance compared to the larger model.
For our experiments with the 2B model, we employed the DDIM sampler, which is also used in the 2B version of CogVideoX.
We compared the performance differences between the 2B and 5B models within our method, specifically focusing on the single-task segmentation model, as presented in Tab.~\ref{tab:supp_model_size}. 
The increased number of parameters in the 5B model led to significant performance improvements. 
This larger model is better at incorporating segmentation guidance and can generate more realistic and higher-quality videos.

\section{More Visualization Results}
\begin{figure*}[tp]
    \centering
    \includegraphics[width=0.99\textwidth]{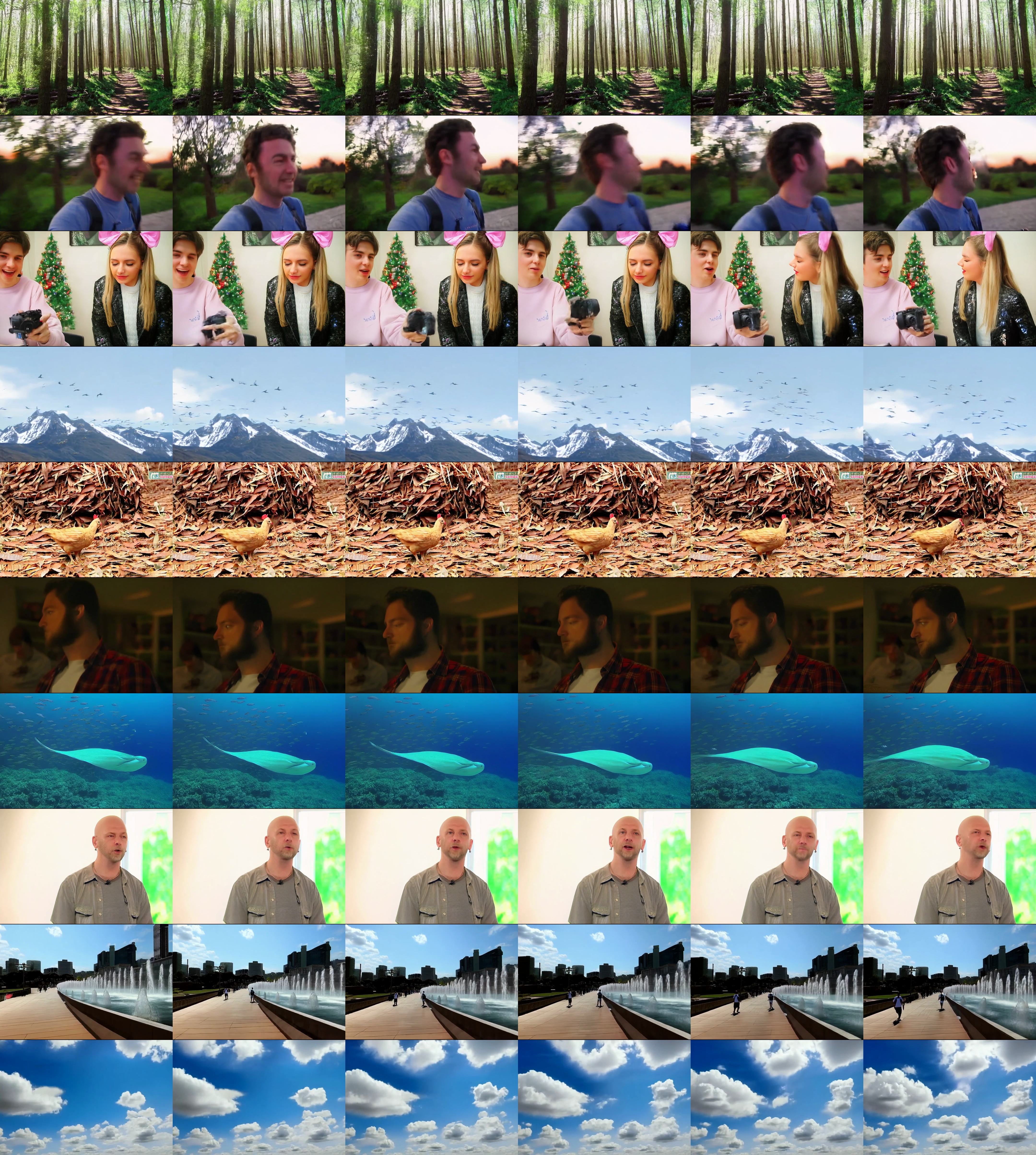}
    \caption{\textbf{More examples on video generation quality}. Six frames are evenly sampled from the entire 49-frame video generated. Having the sample in both portraits, animals, and landscapes.}
    \label{fig.supp_quality}
\end{figure*}

Fig.~\ref{fig.supp_quality} presents additional videos generated by our model, showcasing its capabilities in diverse scenes. 
These videos include subjects such as portraits of people, animals in motion, and dynamic landscapes. 
In these examples, we demonstrate high-quality generation results, producing videos that are consistent, smooth, and aesthetically pleasing. 
The model effectively maintains temporal coherence and visual fidelity throughout the frames, capturing fine details and natural movements inherent in different scenes.

\begin{figure*}[tp]
    \centering
    \includegraphics[width=0.99\textwidth]{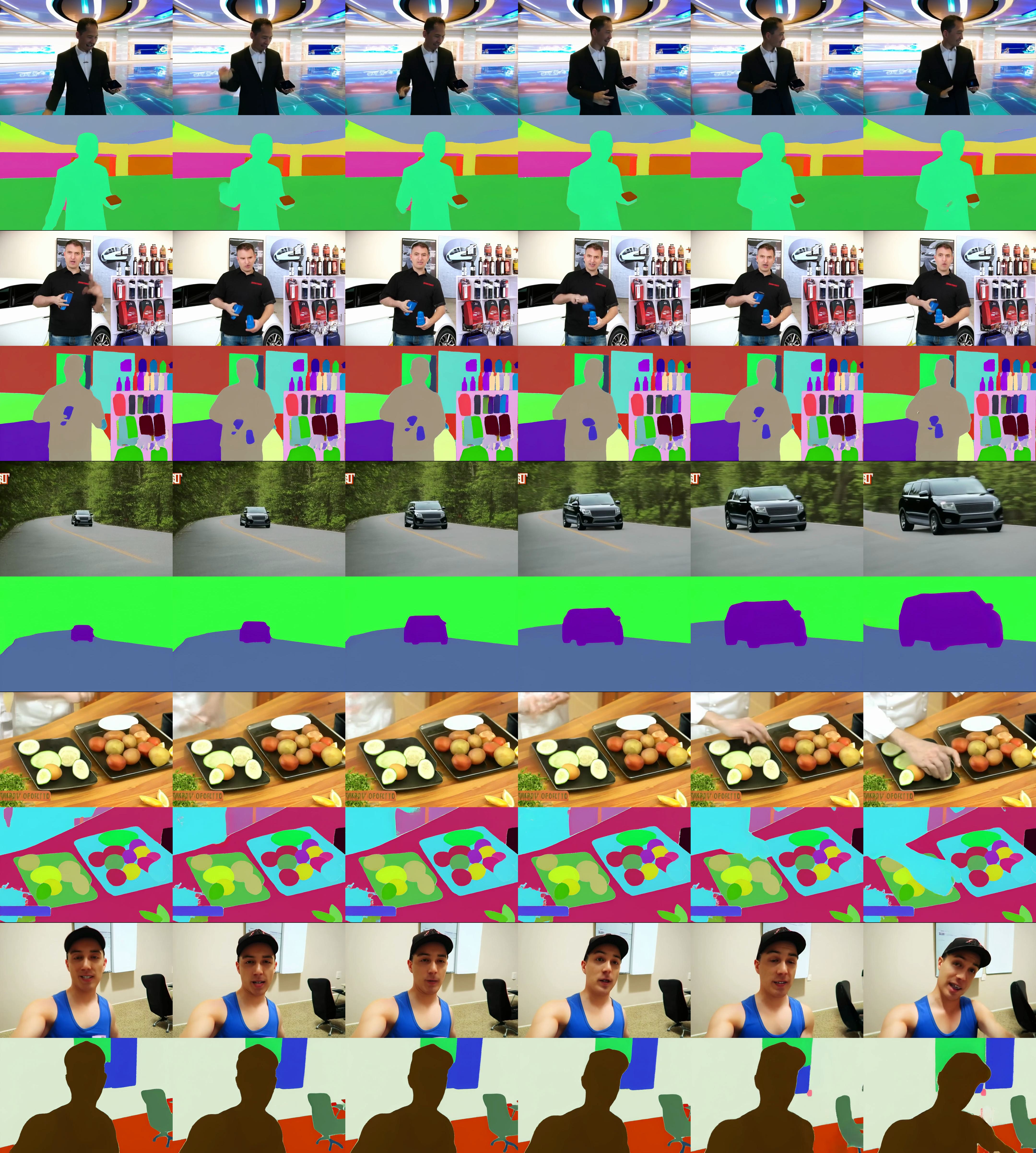}
    \caption{\textbf{More examples on video generation with segmentation}. Each pair consists of two rows: the first row displays the generated video, while the second row shows the corresponding segmentation. Six frames are evenly sampled from the total of 49 frames in the generated video. Our results demonstrate excellent performance in both simple scenes and those with densely packed objects, highlighting our ability to effectively segment dense entities.}
    \label{fig.supp_seg}
\end{figure*}

\begin{figure*}[tp]
    \centering
    \includegraphics[width=0.99\textwidth]{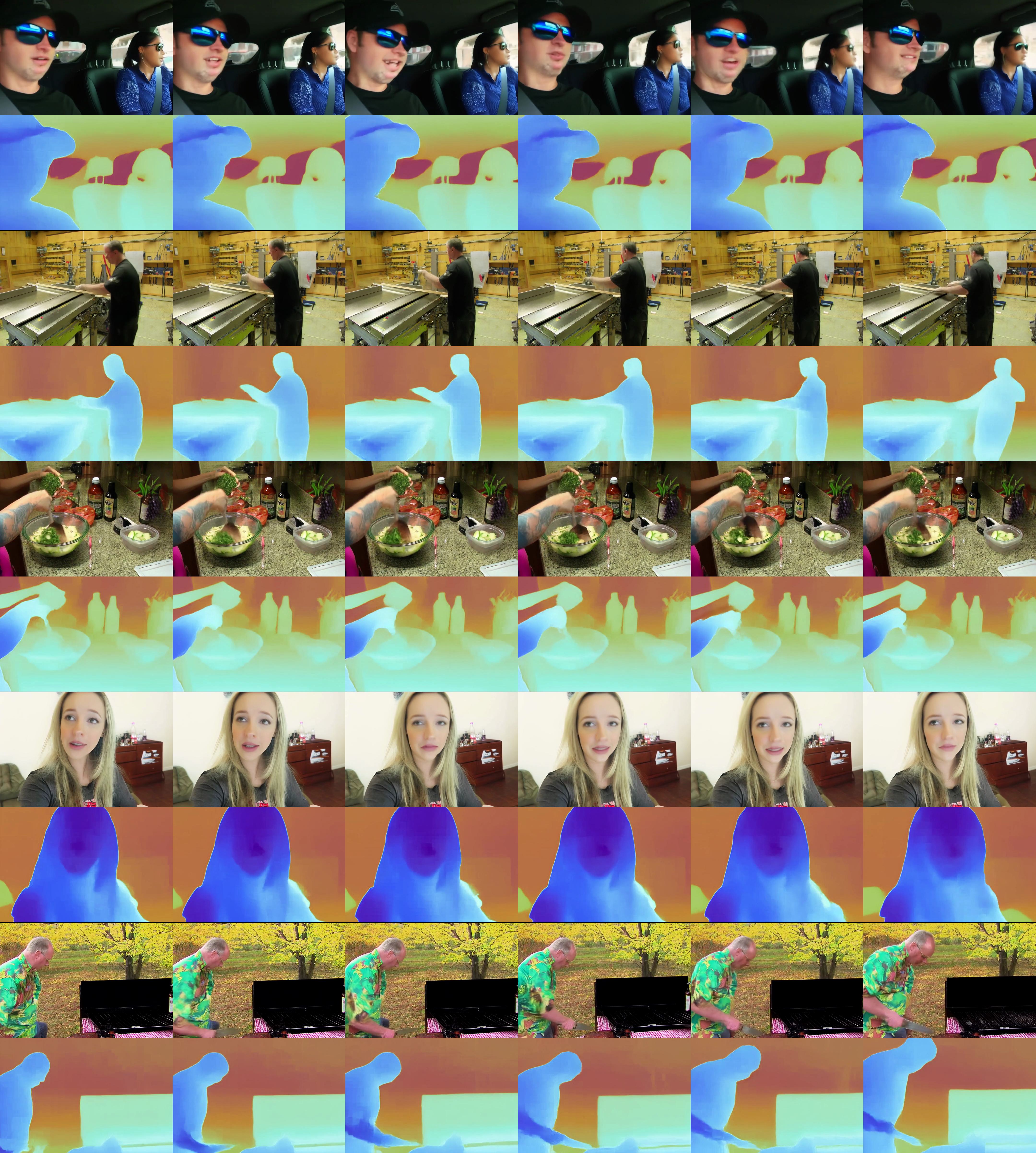}
    \caption{\textbf{More examples on video generation with depth estimation}. Each pair has two rows: the first row displaying the generated video, and the second row presenting the corresponding depth estimation map. Six frames are evenly sampled from the total of 49 frames in the generated video.}
    \label{fig.supp_depth}
\end{figure*}

Fig.~\ref{fig.supp_seg} provides more videos and their segmentation results, where the multitask model performs inference in segmentation mode. 
The guidance provided by segmentation improves the consistency and quality of video generation and produces the corresponding high-quality segmentation masks. 
Our segmentation is very detailed in both simple and complex, densely populated scenes. 
Fig.~\ref{fig.supp_depth} displays videos and their corresponding depth maps, with inference performed in depth estimation mode by the multitask model. 
Depth estimation also improves consistency and quality while generating accurate depth maps.

\end{document}